\documentclass{article} 
\usepackage{times}
\usepackage{amsmath}
\usepackage{amsfonts}
\usepackage{amssymb}
\usepackage[latin1]{inputenc}
\usepackage{graphicx}
\usepackage{epstopdf}
\usepackage{graphicx}
\usepackage{subfigure}
\usepackage{tikz}%
\usepackage{xr}
\usepackage[a4paper]{geometry}
\usepackage{RR}%
\graphicspath{{./}{./figures/}}
\newtheorem{theorem}{Theorem}

\newtheorem{proposition}[theorem]{Proposition}

\graphicspath{{./}{./figures/}}

\def\XX{\mathbb{X}}
\def\RR{\mathbb{R}}
\def\EE{\mathbb{E}}

\def\bone{{\rm 1\hspace*{-0.4ex}%
\rule{0.1ex}{1.52ex}\hspace*{0.2ex}}}

\DeclareMathOperator{\Exp}{Exp}
\DeclareMathOperator{\Poi}{Poisson}
\DeclareMathOperator{\Gam}{Gamma}

\RRNo{8140}
\RRdate{November 19, 2012}
\RRauthor{
Fran\c cois Caron\thanks{INRIA Bordeaux Sud-Ouest, Institut de Math\'ematiques de Bordeaux, University of Bordeaux, France}%
  \and
Yee Whye Teh\thanks{Department of Statistics, Oxford University, United Kingdom}%
}
\authorhead{F. Caron and Y.W. Teh}
\RRtitle{Modèles bayésiens non paramétriques pour les données de rang}
\RRetitle{Bayesian nonparametric models for ranked data}
\titlehead{Bayesian nonparametric models for ranked data}
\RRresume{On s'intéresse dans ce rapport à une extension bayésienne non paramétrique du modèle de Plackett-Luce pour les données de rang, pouvant traiter un nombre potentiellement infini d'éléments. Notre cadre se base sur la théorie des mesures complètement aléatoires, avec comme a priori un processus de gamma. Nous dérivons une caractérisation de la loi a posteriori et un échantillonneur de Gibbs simple pour approcher la loi a posteriori. Nous développons également une version dynamique de notre modèle, et l'appliquons aux listes hebdomadaires des 20 meilleures ventes du New York Times.}
\RRabstract{We develop a Bayesian nonparametric extension of the popular Plackett-Luce choice model that can handle an infinite number of choice items. Our framework is based on the theory of random atomic measures, with the prior specified by a gamma process.  We derive a posterior characterization and a simple and effective Gibbs sampler for posterior simulation.  We develop a time-varying extension of our model, and apply it to the New York Times lists of weekly bestselling books.}
\RRmotcle{Modèles de choix, modèle de Bradley-Terry généralisé, modèle de Plackett-Luce, processus de gamma, méthodes de Monte Carlo par chaîne de Markov}
\RRkeyword{choice models, generalized Bradley-Terry model, Plackett-Luce model, gamma process, Markov Chain Monte Carlo}
\RRprojet{ALEA}

\RCBordeaux 

\begin{document}

\makeRR   



%

\section{Introduction}

Data in the form of partial rankings, i.e.\ in terms of an ordered list of the
top-$m$ items, arise in many contexts.
For example, in this paper we consider datasets consisting of
the top 20 bestselling books as published each week by the New York Times.
The Plackett-Luce model~\cite{Luce1959,Plackett1975} is
a popular model for modeling such partial rankings of a finite collection of $M$ items.
It has found many applications,
including choice modeling~\cite{Luce1977}, sport
ranking~\cite{Hunter2004}, and voting~\cite{Gormley2008a}.
\cite[Chap.\ 9]{Diaconis1988}~\ provides detailed discussions on the
statistical foundations of this model.

In the Plackett-Luce model,
each item $k\in[M]=\{1,\ldots,M\}$ is assigned a positive rating parameter $w_{k}$, which represents the desirability or rating of a product in the case of choice modeling,
or the skill of a player in sport rankings.
The Plackett-Luce model assumes the following generative story for a top-$m$ list
$\rho=(\rho_{1},\ldots,\rho_{m})$ of items $\rho_i\in[M]$:
At each stage $i=1,\ldots,m$,  an item is chosen to be the $i$th
item in the list from among the items that have not yet
appeared, with the probability that $\rho_i$ is selected  being proportional to its desirability $w_{\rho_i}$.
The overall probability of a given partial ranking $\rho$
 is then:
\begin{equation}
P(\rho)=\prod_{i=1}^{m}\frac{w_{\rho_{i}}}{\big(  \sum_{k=1}^{M}%
w_{k}\big)  -\big(  \sum_{j=1}^{i-1}w_{\rho_{j}}\big)  }. \label{eq:finitepl}
\end{equation}
with the denominator in \eqref{eq:finitepl} being the sum over all items not yet selected at stage $i$.

In many situations the collection of available items can be very large and potentially unknown.
In this case, a nonparametric approach can be sensible,
where the pool of items is assumed to be infinite and the model allows for the possibility of items not
observed in previous top-$m$ lists to appear in new ones.  In this paper we propose such a Bayesian nonparametric Plackett-Luce model.  Our approach is built upon recent work on Bayesian inference for the (finite) Plackett-Luce model and its extensions \cite{Gormley2009,Guiver2009,Caron2012}.  Our model assumes the existence of an infinite pool of items $\{X_k\}_{k=1}^\infty$, each with its own rating parameter, $\{w_k\}_{k=1}^\infty$.  The probability of a top-$m$ list of items, say $(X_{\rho_1},\ldots,X_{\rho_m})$, is then a direct extension of the finite case \eqref{eq:finitepl}:
\begin{equation}
P(X_{\rho_1},\ldots,X_{\rho_m})=\prod_{i=1}^{m}\frac{w_{\rho_{i}}}{\big(  \sum_{k=1}^{\infty}%
w_{k}\big)  -\big(  \sum_{j=1}^{i-1}w_{\rho_{j}}\big)  }. \label{eq:infinitepl}
\end{equation}
To formalize the framework, a natural representation to encapsulate the pool of items along with their ratings is using an atomic measure:
\begin{equation}
G = \sum_{k=1}^\infty w_k \delta_{X_k} \label{eq:G}
\end{equation}
Using this representation, note that the top item $X_{\rho_1}$ in our list is simply a draw from the probability measure obtained by normalizing $G$, while subsequent items in the top-$m$ list are draws from probability measures obtained by first removing from $G$ the atoms corresponding to previously picked items and normalizing.  Described this way, it is clear that the Plackett-Luce model is basically a partial size-biased permutation of the atoms in $G$ \cite{PatTai1977a}, and the existing machinery of random measures and exchangeable random partitions \cite{Pitman2006} can be brought to bear on our problem.

In particular, in Section \ref{model} we will use a gamma process as the prior over the
atomic measure $G$.  This is a completely random measure \cite{Kin1967a} with gamma
marginals, such that the corresponding normalized probability measure is a Dirichlet
process.   We will show that with the introduction of a suitable set of auxiliary variables, we
can characterize the posterior law of $G$ given observations of top-$m$ lists distributed
according to \eqref{eq:infinitepl}.  A simple Gibbs sampler can then be derived to simulate
from the posterior distribution.
In Section \ref{sec:dynamic} we develop a time-varying extension of our model
and derive a simple and effective Gibbs sampler for posterior simulation.
In Section \ref{sec:experiments} we apply our time-varying Bayesian nonparametric Plackett-
Luce model to the aforementioned New York Times bestsellers datasets, and conclude in
Section \ref{sec:discussion}.

\section{A Bayesian nonparametric model for partial ranking}\label{model}

We start this section by briefly describing a Bayesian approach to inference in finite Plackett-Luce models \cite{Caron2012}, and taking the infinite limit to arrive at the nonparametric model.  This will give good intuitions for how the model operates, before we rederive the same nonparametric model more formally using gamma processes.  Throughout this paper we will suppose that our data consists of $L$ partial rankings, with $\rho_{\ell}=(\rho_{\ell1},\ldots,\rho_{\ell m})$ for $\ell\in[L]$.  For notational simplicity we assume that all the partial rankings are length $m$.

\subsection{Finite Plackett-Luce model with gamma prior}

Suppose we have $M$ choice items, with item $k\in[M]$ having a positive desirability parameter $w_k$.  A partial ranking $\rho_{\ell}=(\rho_{\ell1},\ldots,\rho_{\ell m})$ can be constructed generatively by picking the $i$th item $\rho_{\ell i}$ at the $i$th stage for $i=1,\ldots,m$, with probability proportional to $w_{\rho_{\ell i}}$ as in \eqref{eq:finitepl}.  An alternative Thurstonian interpretation, which will be important in the following, is as follows:  For each item $k$ let $z_{\ell k}\sim\Exp(w_k)$ be exponentially distributed with rate $w_k$. Thinking of $z_{\ell k}$ as the arrival time of item $k$ in a race, let $\rho_{\ell i}$ be the index of the $i$th item to arrive (the $i$th smallest value among $(z_{\ell k})_{k=1}^M$).  The resulting probability of $\rho_\ell$ can then be shown to still be \eqref{eq:finitepl}.  In this interpretation $(z_{\ell k})$ can be understood as latent variables, and the EM algorithm can be applied to derive an algorithm to find a ML parameter setting for $(w_k)_{k=1}^M$ given multiple partial rankings.  Unfortunately the posterior distribution of $(z_{\ell k})$ given $\rho_\ell$ is difficult to compute directly, so we instead consider an alternative parameterization: Let $Z_{\ell i}=z_{\rho_{\ell i}}-z_{\rho_{\ell\, i-1}}$ be the waiting time for the $i$th item to arrive after the $i-1$th item (with $z_{\rho_{\ell 0}}$ defined to be 0).  Then it can be shown that the joint probability is:
\begin{align}
P((\rho_\ell)_{\ell=1}^L,(Z_{\ell i})_{\ell=1,i=1}^{L,m}|(w_k)_{k=1}^M) = \prod_{\ell=1}^L \prod_{i=1}^m w_{\rho_{\ell i}}\exp\left(-Z_{\ell i}\left(\textstyle \sum_{k=1}^M w_k - \sum_{j=1}^{i-1} w_{\rho_{\ell j}}\right)\right) \label{eq:finitejoint}
\end{align}
Note that the posterior of $(Z_{\ell i})_{i=1}^m$ is simply factorized with $Z_{\ell i}|\rho,w\sim\Exp(\sum_{k=1}^M w_k - \sum_{j=1}^{i-1} w_{\rho_{\ell j}})$, and the ML parameter setting can be easily derived as well.  Taking a further step, we note that  a factorized gamma prior over $(w_k)$ is conjugate to \eqref{eq:finitejoint}, say $w_k\sim\Gam(\frac{\alpha}{M},\tau)$ with hyperparameters $\alpha,\tau>0$.  Now Bayesian inference can be carried out either with a VB EM algorithm, or a Gibbs sampler.  In this paper we shall consider only Gibbs sampling algorithms.  In this case the parameter updates are of the form
\begin{align}
w_k|(\rho_\ell),(Z_{\ell i}),(w_{k'})_{k'\neq k} &\sim \Gam\textstyle\left(
 \frac{\alpha}{M}+n_k,
\tau+\sum_{\ell=1}^L\sum_{i=1}^m \delta_{\ell ik} Z_{\ell i}
\right) \label{eq:finiteposterior}
\end{align}
where $n_k$ is the number of occurrences of item $k$ among the observed partial rankings, and $\delta_{\ell ik}=0$ if there is a $j< i$ with $\rho_{\ell j}=k$ and 1 otherwise.  These terms arise by regrouping those in the exponential in \eqref{eq:finitejoint}.

A nonparametric Plackett-Luce model can now be easily derived by taking the limit as the number of choice items $M\rightarrow\infty$.  For those items $k$ that have appeared among the observed partial rankings, the limiting conditional distribution \eqref{eq:finiteposterior} is well defined since $n_k>0$.  For items that did not appear in the observations, \eqref{eq:finiteposterior} becomes degenerate at 0.  Instead we can define $w_*=\sum_{k:n_k=0} w_k$ to be the total desirability among all infinitely many previously unobserved items, and show that
\begin{align}
w_*|(\rho_\ell),(Z_{\ell i}),(w_{k})_{k:n_k>0} &\sim \Gam\textstyle\left(
 \alpha,
\tau+\sum_{\ell=1}^L\sum_{i=1}^m Z_{\ell i}
\right)
\end{align}
The Gibbs sampler thus alternates between updating $(Z_{\ell i})$, and updating the ratings of the observed items $(w_k)_{k:n_k>0}$ and of the unobserved ones $w_*$.  This nonparametric model allows us to estimate the probability of seeing new items appearing in future partial rankings in a consistent manner.   While intuitive, this derivation is ad hoc in the sense that it arises as the infinite limit of the Gibbs sampler for finite models, and is unsatisfying as it did not directly capture the structure of the underlying infinite dimensional object, which we will show in the next subsection to be a gamma process.

\subsection{A Bayesian nonparametric Plackett-Luce model}\label{crmpl}

Let $\XX$ be a measurable space of choice items.  A gamma process is a completely random measure over $\XX$ with gamma marginals.  Specifically, it is a random atomic measure of the form \eqref{eq:G}, such that for each measurable subset $A$, the (random) mass $G(A)$ is gamma distributed.  Assuming that $G$ has no fixed atoms (that is, for each element $x\in \XX$ we have $G(\{x\})=0$ with probability one) and that the atom locations $\{X_k\}$ are independent of their masses $\{w_k\}$, it can be shown that such a random measure can be constructed as follows: each $X_k$ is iid according to a base distribution $H$ (which we assume is non-atomic with density $h(x)$), while the set of masses $\{w_k\}$ is distributed according to a Poisson process over $\RR^+$ with intensity $\lambda(w) = \alpha w^{-1}e^{-w\tau}$ where
 $\alpha>0$ is the concentration parameter and $\tau>0$ the inverse
scale.  We write this as $G\sim\Gamma(\alpha,\tau,H)$.  Under this parametrization, we have that $G(A)\sim\Gam
(\alpha H(A),\tau)$.

\begin{figure}
\begin{minipage}{.38\textwidth}
{\includegraphics[width=\textwidth]{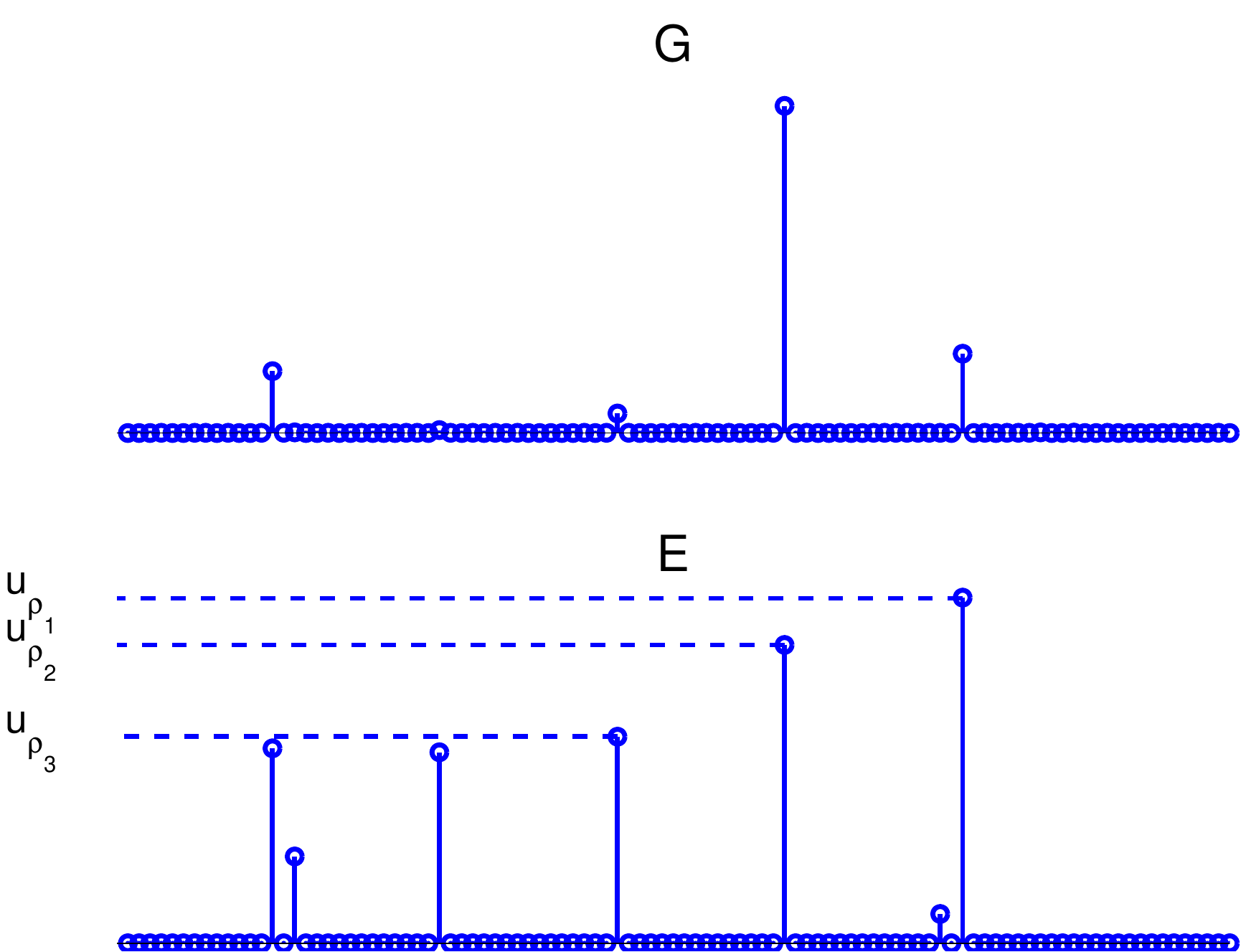}}
\end{minipage}
\begin{minipage}{.6\textwidth}
{\includegraphics[width=.32\textwidth]{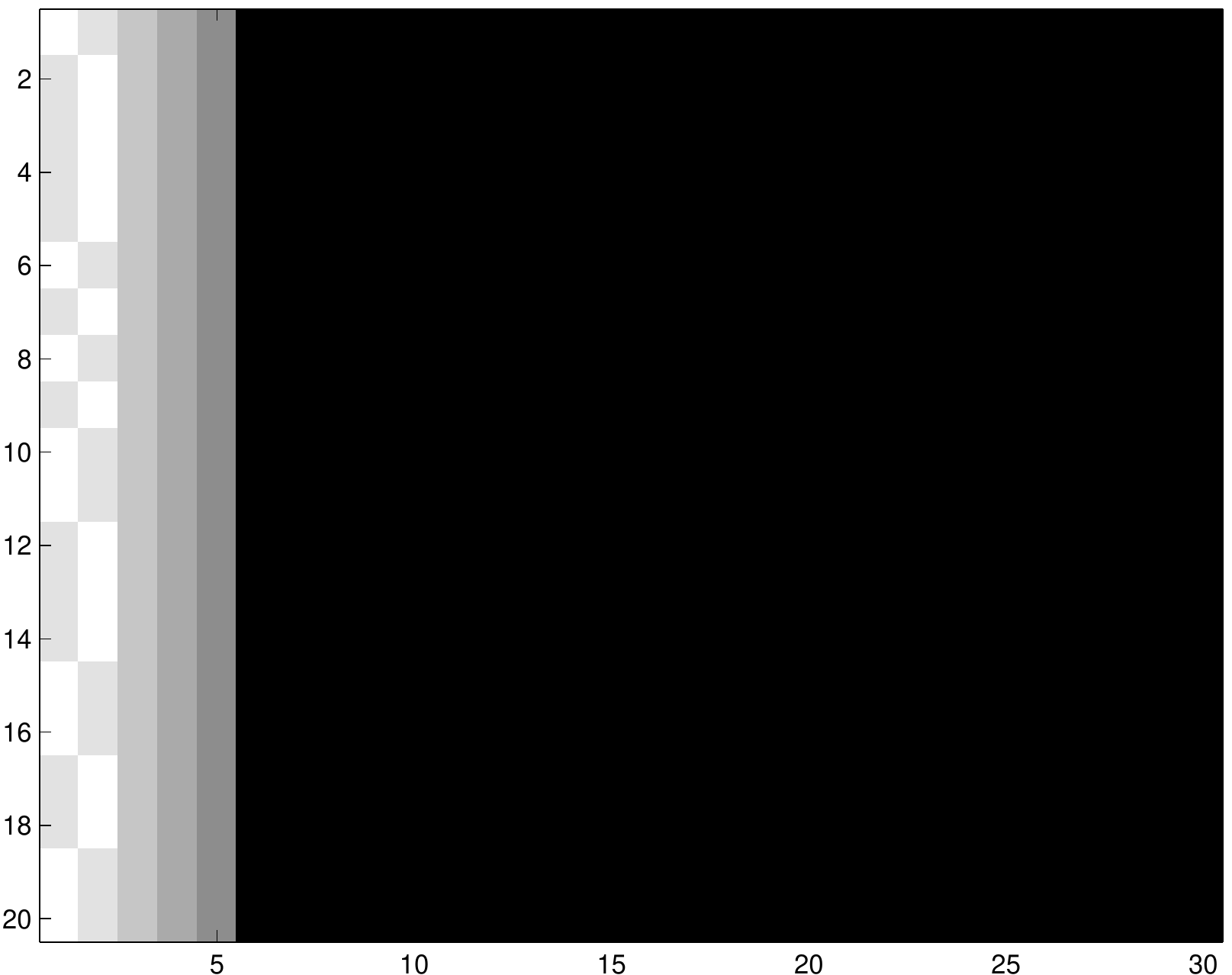}}
{\includegraphics[width=.32\textwidth]{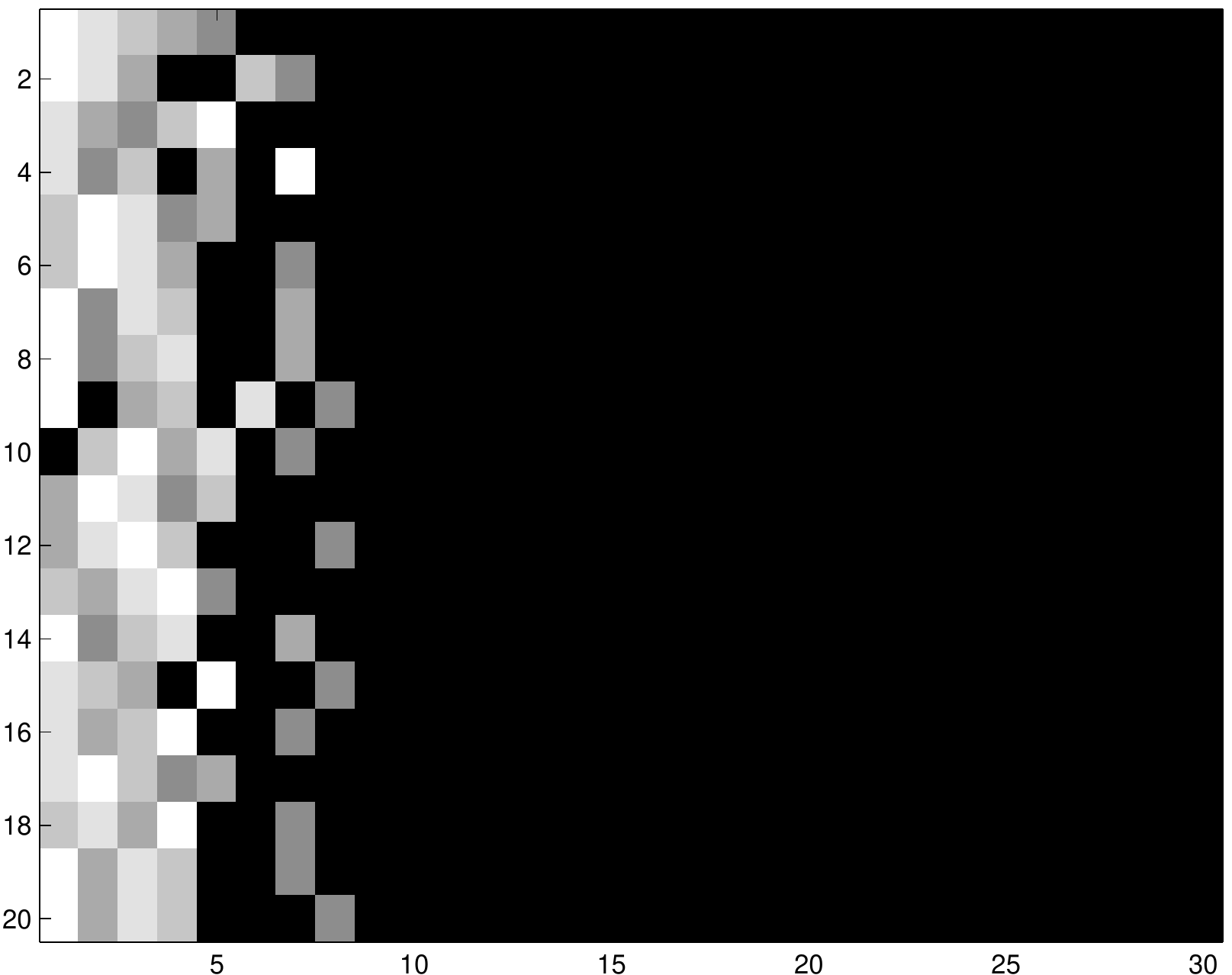}}
{\includegraphics[width=.32\textwidth]{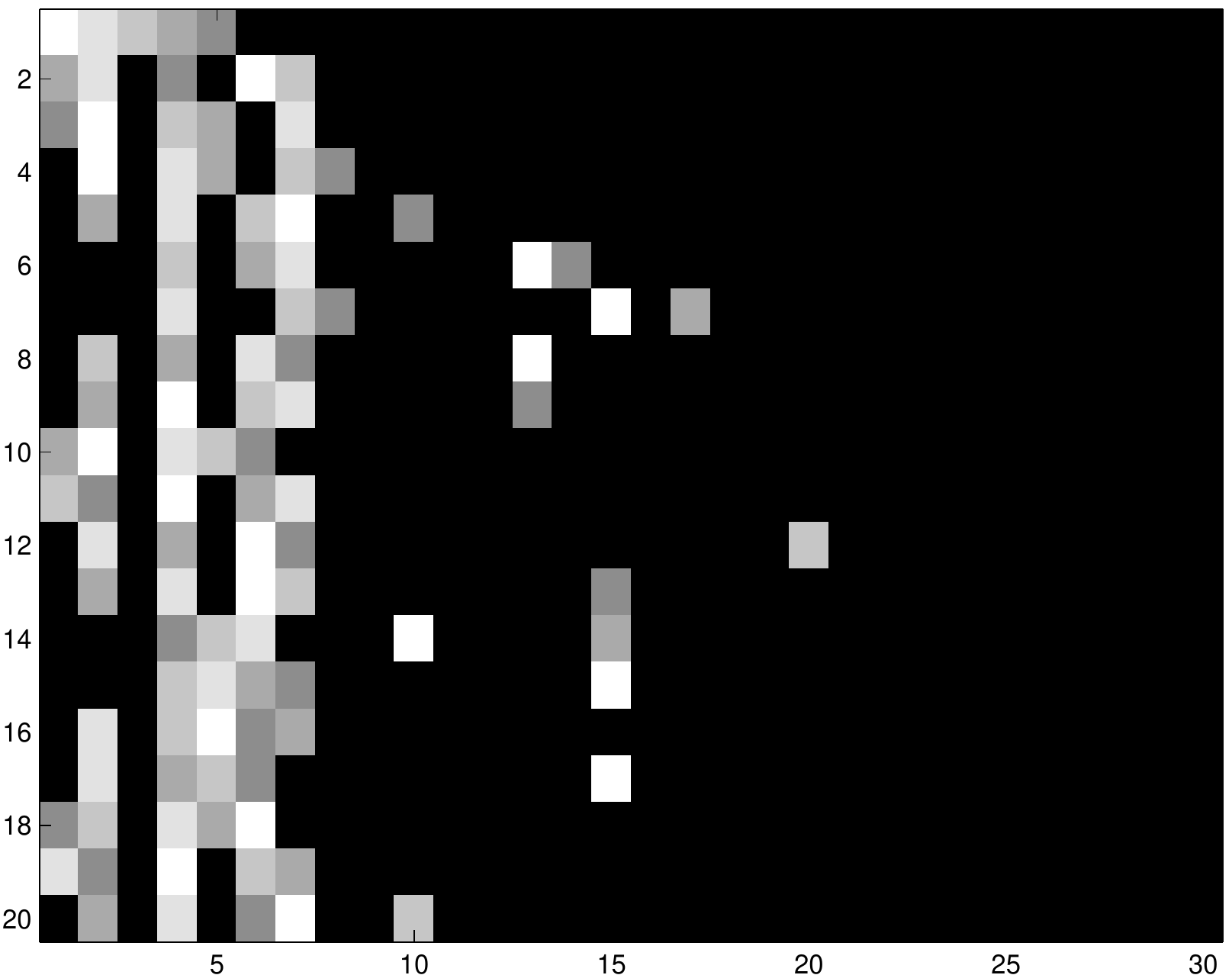}}
\caption{Bayesian nonparametric Plackett-Luce model.
Left: $G$ and $U=\sum_k u_k\delta_{X_k}$ where $u_k=-\log(z_k)$. The top-3 ranking is $(\rho_1,\rho_2,\rho_3)$.   Right:  Visualization of top-5 rankings with rows corresponding to different rankings and columns to items sorted by size biased order. A lighter shade corresponds to a higher rank.  Each figure is for a different $G$, with $\alpha=.1,1,3$. }%
\label{fig:CRM}%
\end{minipage}
\end{figure}

Each atom $X_k$ is a choice item, with its mass $w_{k}>0$
corresponding to the desirability parameter.  The
Thurstonian view described in the finite model can be easily extended
to the nonparametric one, where a partial ranking $(X_
{\rho_{\ell 1}} \ldots X_{\rho_{\ell m}})$ can be generated as the first $m$ items
to arrive in a race.  In particular, for each atom $X_k$ let $z_{\ell k}\sim
\Exp(w_k)$ be the time of arrival of
$X_k$ and  $X_{\rho_{\ell i}}$  the $i$th item to arrive. The first $m$
items to arrive $(X_{\rho_{\ell 1}} \ldots X_{\rho_{\ell m}})$ then constitutes our top-$m$ list, with probability as given in \eqref{eq:infinitepl}.  Again
reparametrizing using inter-arrival durations, let $Z_{\ell i}=z_{\rho_{\ell i}}-z_
{\rho_{\ell i-1}}$ for $i=1,2,\ldots$ (with $z_{\rho_0}=0$).  Then the joint
probability is:
\begin{align}
&P((X_{\rho_{\ell i}})_{i=1}^m,(Z_{\ell i})_{i=1}^m|G)
= P((z_{\rho_{\ell 1}} \ldots z_{\rho_{\ell m}}),\text{and $z_{\ell k}>z_{\rho_{\ell m}}$ for all $k\not\in\{\rho_{\ell 1},\ldots,\rho_{\ell m}\}$}) \label{eq:xandz}
\\
=&\bigg(\prod_{i=1}^m w_{\rho_{\ell i}}e^{-w_{\rho_{\ell i}}z_{\rho_{\ell i}}}\bigg)\bigg(\prod_{k\not\in\{\rho_{\ell i}\}_{i=1}^m} \hspace*{-1em}e^{-w_k z_{\rho_{\ell m}}}\bigg)
=\prod_{i=1}^m w_{\rho_{\ell i}}\exp\hspace*{-.1em}\bigg(\hspace*{-.3em}-\hspace*{-.1em}Z_{\ell i}\bigg(\sum_{k=1}^\infty w_k-\sum_{j=1}^{i-1}w_{\rho_{\ell j}}\bigg)\hspace*{-.3em}\bigg)\nonumber
\end{align}
Marginalizing out $(Z_{\ell i})_{i=1}^m$ gives the probability of $(X_{\rho_{\ell i}})_{i=1}^m$ in \eqref{eq:infinitepl}.  Further, conditional on $\rho_\ell$ it is seen that the inter-arrival durations $Z_{\ell 1} \ldots Z_{\ell m}$ are mutually independent and exponentially distributed:
\begin{align}
Z_{\ell i}|(X_{\rho_{\ell i}})_{i=1}^m,G  &  \sim\Exp\bigg(\sum_{k=1}^{\infty}w_{k}-\sum_{j=1}^{i-1}w_{\rho_{\ell j}}\bigg)
\end{align}
The above construction is depicted on Figure~\ref{fig:CRM}(left).  We visualize on right some top-$m$ lists generated from the model, with $\tau=1$ and different values of $\alpha$.

\subsection{Posterior characterization}\label{posterior}

Consider a number $L$ of partial rankings, with the $\ell$th list denoted $Y_\ell=(Y_{\ell1} \ldots Y_{\ell m_\ell})$ , for $\ell\in[L]$.  While previously our top-$m$ list $(X_{\rho_1}\ldots X_{\rho_m})$ consists of an ordered list of the atoms in $G$.  Here $G$ is unobserved and $(Y_{\ell 1}\ldots Y_{\ell m_\ell})$ is simply a list of observed choice items, which is why they were not expressed as an ordered list of atoms in $G$.  The task here is then to characterize the posterior law of $G$ under a gamma process prior and supposing that the observed partial rankings were drawn iid from the nonparametric Plackett-Luce model given $G$.  Re-expressing the conditional distribution \eqref{eq:infinitepl} of $Y_\ell$ given $G$, we have:
\begin{align}
P(Y_\ell|G) = \prod_{i=1}^{m_\ell} \frac{G(\{Y_{\ell i}\})}{G(\XX\backslash\{Y_{\ell 1} \ldots Y_{\ell\, i-1}\})}
\end{align}
As before, for each $\ell$, we will also introduce a set of auxiliary variables $Z_\ell=(Z_{\ell 1} \ldots Z_{\ell m_\ell})$ (the inter-arrival times) that are conditionally mutually independent given $G$ and $Y_\ell$, with:
\begin{align}
Z_{\ell i}|Y_\ell, G &\sim \Exp(G(\mathbb{X}\backslash\{Y_{\ell 1},\ldots,Y_{\ell i-1}\}))\label{eq:condz}
\end{align}
The joint probability of the item lists and auxiliary variables is then (c.f.\ \eqref{eq:xandz}):
\begin{align}
P((Y_\ell,Z_\ell)_{\ell=1}^L|G) = \prod_{\ell=1}^L \prod_{i=1}^{m_\ell}G(\{Y_{\ell i}\})\exp(-Z_{\ell i}G(\mathbb{X}\backslash\{Y_{\ell 1},\ldots,Y_{\ell\, i-1}\})) \label{eq:jointG}
\end{align}
Note that under the generative process described in Section~\ref{crmpl}, there is positive probability that an item appearing in a list $Y_\ell$ appears in another list $Y_{\ell'}$ with $\ell'\neq \ell$.  Denote the unique items among all $L$ lists by $X^*_1 \ldots X^*_K$, and for each $k=1,\ldots,K$ let $n_k$ be the number of occurrences of $X^*_k$ among the item lists.  Finally define occurrence indicators
\begin{align}
\delta_{\ell i k} &= \begin{cases}
0 & \text{if $\exists j< i$ with $Y_{\ell j}=X^*_k$;} \\
1 & \text{otherwise.}
\end{cases}
\end{align}
i.e. $\delta_{\ell i k}$ is the indicator of the occurence that item $X^*_k$ does not appear at a rank lower than $i$ in the $\ell$th list. Then the joint probability under the nonparametric Plackett-Luce model is:
\begin{align}
P((Y_\ell,Z_\ell)_{\ell=1}^L|G) &=
\prod_{k=1}^K G(\{X^*_{k}\})^{n_k}\times
\prod_{\ell=1}^L \prod_{i=1}^{m_\ell}\exp(-Z_{\ell i}
G(\mathbb{X}\backslash\{Y_{\ell 1},\ldots,Y_{\ell\, i-1}\})) \nonumber \\
&= \exp\left(-G(\mathbb{X})\sum_{\ell i}Z_{\ell i}\right)
\prod_{k=1}^K G(\{X^*_{k}\})^{n_k}\exp\left(-G(\{X^*_k\})\sum_{\ell i}(\delta_{\ell ik}-1)Z_{\ell i}\right)
\label{eq:likelihood}
\end{align}
Taking expectation of \eqref{eq:likelihood} with respect to $G$ using the Palm formula gives:

\begin{theorem} \label{thm:marginal}
The marginal probability of the $L$ partial rankings and auxiliary variables is:
\begin{align}
P((Y_\ell,Z_\ell)_{\ell=1}^L) =
e^{-\psi(\sum_{\ell i}Z_{\ell i})} \prod_{k=1}^K h(X^*_k) \kappa\bigg(n_k,\sum_{\ell i} \delta_{\ell ik} Z_{\ell i}\bigg)\label{eq:joint}
\end{align}
where $\psi(z)$ is the Laplace transform of $\lambda$,
\begin{align}
\psi(z) = -\log\EE\left[e^{-z G(\XX)}\right] = \int_{\RR^+} \lambda(w) (1-e^{-zw}) dw=\alpha\log\left(1+\frac{z}{\tau}\right)
\end{align}
and $\kappa(n,z)$ is the $n$th moment of the exponentially tilted L\'evy intensity $\lambda(w)e^{-zw}$:
\begin{align}
\kappa(n,z) = \int_{\RR^+} \lambda(w) w^n e^{-zw} dw=\frac{\alpha}{(z+\tau)^n}\Gamma(n)
\end{align}
\end{theorem}


Details are given in the appendix.  Another application of the Palm formula now allows us to derive a posterior characterisation of $G$:

\begin{theorem} \label{thm:posterior}
Given the observations and associated auxiliary variables $(Y_\ell,Z_\ell)_{\ell=1}^L$,  the posterior law of $G$ is also a gamma process, but with atoms with both fixed and random locations. Specifically,
\begin{align}
G|(Y_\ell,Z_\ell)_{\ell=1}^L &= G^* + \sum_{k=1}^K w^*_k \delta_{X^*_k} \label{eq:posterior_rep}
\end{align}
where $G^*$ and $w^*_1,\ldots,w^*_K$ are mutually independent.  The law of $G^*$ is still a gamma process,
\begin{align}
G^* |(X_\ell,Z_\ell)_{\ell=1}^L &\sim \Gamma(\alpha,\tau^*,h) &
\tau^* &= \tau +\sum_{\ell i} Z_{\ell i}
\end{align}
while the masses have distributions,
\begin{align}
w^*_k|(Y_\ell,Z_\ell)_{\ell=1}^L &\sim \Gam\bigg( n_k,\tau + \sum_{\ell i} \delta_{\ell i k}Z_{\ell i}\bigg)
\end{align}
\end{theorem}

\subsection{Gibbs sampling}
Given the results of the previous section, a simple Gibbs sampler can now be derived, where all the conditionals are of known analytic form.
In particular, we will integrate out all of $G^*$ except for its total mass  $w^*_*=G^*(\mathbb X)$.  This leaves the latent variables to consist of the masses $w^*_*$, $(w^*_k)$ and the auxiliary variables $(Z_{\ell i})$.  The update for $Z_{\ell i}$ is given by  \eqref{eq:condz}, while those for the masses are given in Theorem~\ref{thm:posterior}:
\begin{align}
&\text{Gibbs update for $Z_{\ell i}$:}&
Z_{\ell i}|\text{rest} &\sim \Exp \big(w^*_*+\textstyle \sum_{k} \delta_{\ell ik} w^*_k \big) \\
&\text{Gibbs update for $w^*_k$:}&
w^*_k|\text{rest} &\sim \Gam\big (n_k,  \tau+\textstyle\sum_{\ell i} \delta_{\ell ik} Z_{\ell i}\big ) \\
&\text{Gibbs update for $w^*_*$:}&
w^*_*|\text{rest} &\sim \Gam \big ( \alpha, \tau+\textstyle\sum_{\ell i}Z_{\ell i} \big ) \label{eq:gibbst}
\intertext{Note that the auxiliary variables are conditionally independent given
the masses and vice versa.
Hyperparameters of the gamma process can be simply derived from the joint distribution in Theorem~\ref{thm:marginal}.  Since the marginal probability of the partial rankings is invariant to rescaling of the masses, it is sufficient to keep $\tau$ fixed at 1.  As for $\alpha$, if a $\Gam(a,b)$ prior is placed on it, its conditional distribution is still gamma:}
&\text{Gibbs update for $\alpha$:}&
\alpha|\text{rest} &\sim \Gam \big ( a+K, b+\log\big (1+\textstyle\frac{\sum_{\ell i}Z_{\ell i}}{\tau}\big )\big)
\end{align}
Note that this update was derived with $w^*_*$ marginalized out, so after an update to $\alpha$ it is necessary to immediately update $w^*_*$ via \eqref{eq:gibbst} before proceeding to update other variables.

\section{Dynamic Bayesian nonparametric ranking models}
\label{sec:dynamic}
In this section we develop an extension of the Bayesian nonparametric Plackett-Luce model to model time-varying rankings, where the rating parameters of items may change smoothly over time and reflected in a changing series of rankings.  Given a series of times indexed by $t=1,2,\ldots$, we may model the rankings at time $t$ using a  gamma process distributed random measure $G_t$ as in Section~\ref{crmpl}, with Markov dependence among the sequence of measures $(G_t)$ enabling dependence among the rankings over time.
%

\subsection{Pitt-Walker dependence model}

We will construct a dependent sequence  $(G_t)$ which marginally follow a gamma process $\Gamma(\alpha,\tau,H)$ using the construction of \cite{Pitt2005}.
Suppose $G_t\sim\Gamma(\alpha,\tau,H)$.   Since $G_t$ is atomic, we can write it in the form:
\begin{align}
G_t = \sum_{k=1}^\infty w_{tk} \delta_{X_{tk}}
\end{align}
Define a random measure $C_{t}$ with conditional law:
\begin{align}
C_{t}|G_t &= \sum_{k=1}^\infty c_{tk} \delta_{X_{tk}} &
c_{tk}|G_t &\sim \Poi(\phi_{t}w_{tk}) \label{eq:hst}
\end{align}
where $\phi_{t}>0$ is a dependence parameter.
Using the same method as in Section~\ref{posterior}, we can show:
\begin{proposition}\label{lem:pittwalker}
Suppose the law of $G_t$ is $\Gamma(\alpha,\tau,H)$.  The conditional law of $G_t$ given $C_{t}$ is then:
\begin{align}
G_t = G_t^* + \sum_{k=1}^\infty w^*_{tk} \delta_{X_{tk}}
\end{align}
where $G_t^*$ and $(w^*_{tk})_{k=1}^\infty $ are all mutually independent.  The law of $G_t^*$ is given by a gamma process,
while the masses are conditionally gamma,
\begin{align}
G_t^*|C_{t} &\sim \Gamma(\alpha, \tau+\phi_t,H) &
w^*_{tk}|C_{t} &\sim \Gam(c_{tk}, \tau+\phi_t) \label{pww}
\end{align}
\end{proposition}

The idea of \cite{Pitt2005} is to define the conditional law of $G_{t+1}$ given $G_t$  and $C_{t}$ to coincide with the conditional law of $G_t$ given $C_{t}$ as in Proposition~\ref{lem:pittwalker}.  In other words, define
\begin{align}
G_{t+1} &= G_{t+1}^* + \sum_{k=1}^\infty w_{t+1,k} \delta_{X_{tk}} \label{eq:gs}
\end{align}
where $G_{t+1}^*\sim\Gamma(\alpha,\tau+\phi_t,H)$ and $w_{t+1,k}\sim\Gam(c_{tk},\tau+\phi_t)$ are mutually independent.
If the prior law of $G_t$ is $\Gamma(\alpha,\tau,H)$, the marginal law of $G_{t+1}$ will be $\Gamma(\alpha,\tau,H)$ as well when both $G_t$ and $C_{t}$ are marginalized out, thus maintaining a form of stationarity.  Further, although we have described the process in order of increasing $t$, the joint law of $G_t,C_{t},G_{t+1}$ can equivalently be described in the reverse order with the same conditional laws as above.
Note that if $c_{tk}=0$, the conditional distribution of $w_{t+1,k}$ will be degenerate at 0.  Hence  $G_{t+1}$ has an atom at $X_{tk}$ if and only if $C_{t}$ has an atom at $X_{tk}$, that is, if $c_{tk}>0$.  In addition, it also has atoms (those in $G_{t+1}^*$) where $C_{t}$ does not (nor does $G_t$).
Finally, the parameter $\phi_{t}$ can be interpreted as controlling the strength of dependence between $G_{t+1}$ and $G_t$.  Indeed it can be shown that
\begin{align}
\mathbb E [G_{t+1}|G_t] = \frac{\phi_t}{\phi_t+\tau}G_t + \frac{\tau}{\phi_t + \tau} H.
\end{align}
Another measure of dependence can be gleaned by examining the ``lifetime'' of an atom.  Suppose $X$ is an atom in $G_1$ with mass $w>0$.  The probability that $X$ is an atom in $C_2$ with positive mass is $1-\exp(-\phi_1 w)$, in which case it has positive mass in $G_2$ as well.  Conversely, once it is not an atom, it will never be an atom in the future since the base distribution $H$ is non-atomic.  The lifetime of the atom is then the smallest $t$ such that it is no longer an atom. We can show by induction that: (details in appendix)
\begin{proposition}
\label{proposition4}
The probability that an atom $X$ in $G_1$ with mass $w>0$ is dead at time $t$ is given by
$$
P(G_t(\{X\})=0|w)=\exp(-y_{t|1}w)
$$
where $y_{t|1}$ can be obtained by the recurrence $y_{t|t-1}=\phi_{t-1}$ and
$
y_{t|s-1}=\frac{y_{t|s}\phi_{s-1}}{\phi_{s-1}+\tau+y_{t|s}}
$.
\end{proposition}

\subsection{Posterior characterization and Gibbs sampling}

Assume for simplicity that at each time step $t=1,\ldots,T$ we observe one top-$m$ list $Y_t=(Y_{t1},\ldots,Y_{tm})$ (it trivially extends to multiple partial rankings of differing sizes).  We extend the results of the previous section in characterizing the posterior and developing a Gibbs sampler for the dynamical model.

Since each observed item at time $t$ has to be an atom in its corresponding random measure $G_t$, and atoms in $G_t$ can propagate to neighboring random measures via the Pitt-Walker dependence model, we conclude that the set of all observed items (through all times) has to include all fixed atoms in the posterior of $G_t$.  Thus let $X^*=(X_{k}^*)$, $k=1,\ldots,K$ be the set of unique items observed in $Y_1,\ldots,Y_T$, let $n_{tk}\in\{0,1\}$ be the number of times the item $X^*_k$ appears at time $t$, and let $\rho_t$ be defined as $Y_t=(X^*_{\rho_1},\ldots,X^*_{\rho_m})$.
We write the masses of the fixed atoms as $w_{tk}=G_t(\{X^*_{k}\})$, while the total mass of all other random atoms is denoted $w_{t*}=G_t(\mathbb{X}\backslash X^*)$.   Note that $w_{tk}$ has to be positive on a random contiguous interval of time that includes all observations of $X^*_k$---it's lifetime---but is zero outside of the interval.  We also write $c_{tk}=C_t(\{X^*_{k}\})$ and $c_{t*}=C_t(\mathbb{X}\backslash X^*)$.
As before, we introduce, for $t=1,\ldots,T$ and $i=1,\ldots,m$, latent variables
\begin{align}
Z_{ti} \sim \Exp \bigg(w_{t*} +\sum_{k=1}^K w_{tk} - \sum_{j=1}^{i-1}w_{t\rho_j}\bigg)
\end{align}
Each iteration of the Gibbs sampler then proceeds as follows (details in appendix).
The latent variables $(Z_{ti})$ are updated as above.  Conditioned on the latent variables $(Z_{ti})$, $(c_{tk})$ and $(c_{t*})$, we update the masses $(w_{tk})$, which are independent and gamma distributed since all likelihoods are of gamma form.  Note that the total masses $(G_t(\mathbb X))$ are not likelihood identifiable, so we introduce an extra step  to improve mixing by sampling them from the prior (integrating out $(c_{tk})$, $(c_{t*})$), scaling all masses along with it.  Directly after this step we update $(c_{tk})$, $(c_{t*})$.  We update $\alpha$ along with the random masses $(w_{t*})$ and $(c_{t*})$ efficiently using a forward-backward recursion.  Finally, the dependence parameters $(\phi_t)$ are updated.

\subsection{Continuous time formulation using superprocesses}

\begin{figure}
\begin{center}
\includegraphics[width=.47\textwidth,height=5em]{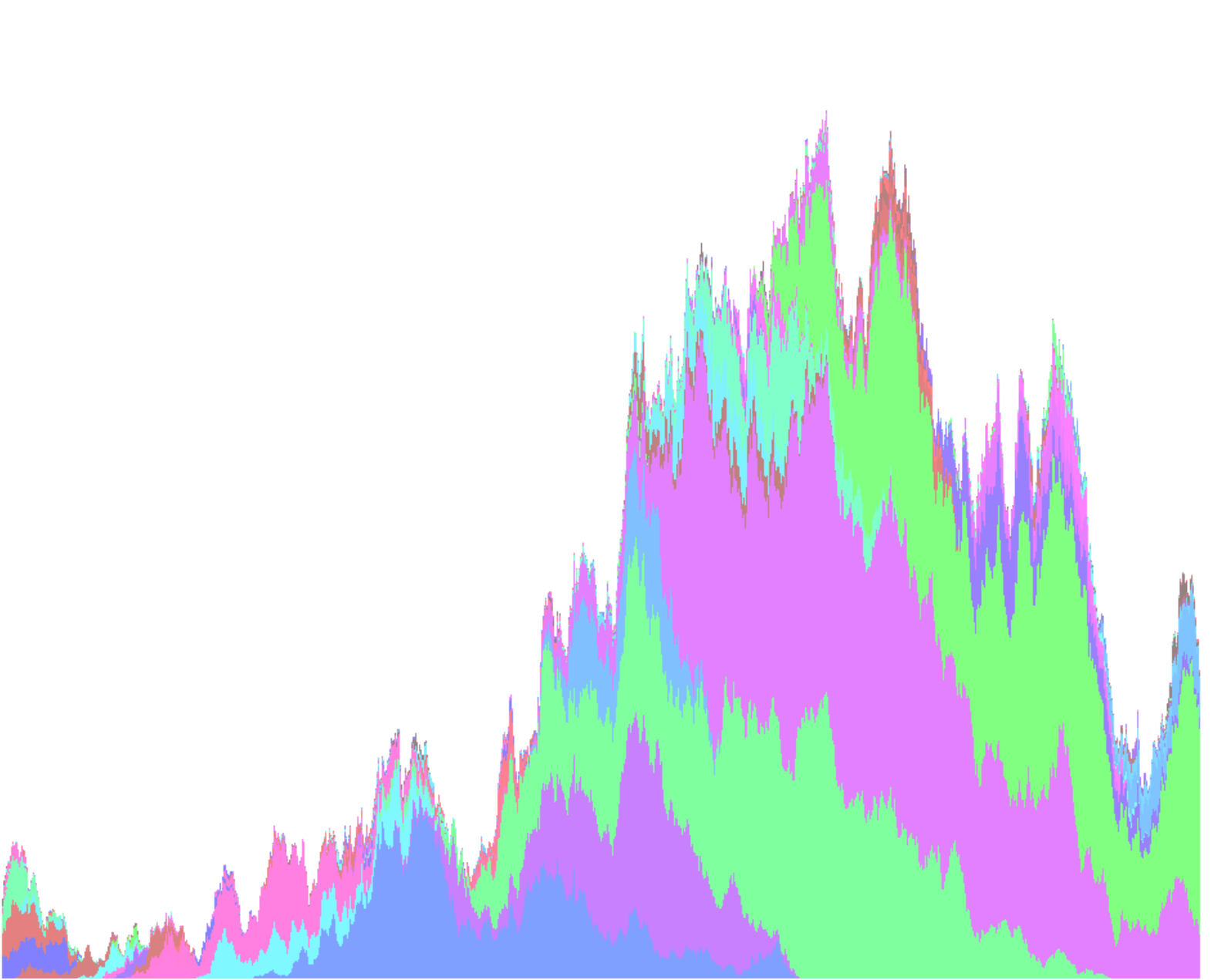}
\includegraphics[width=.47\textwidth,height=5em]{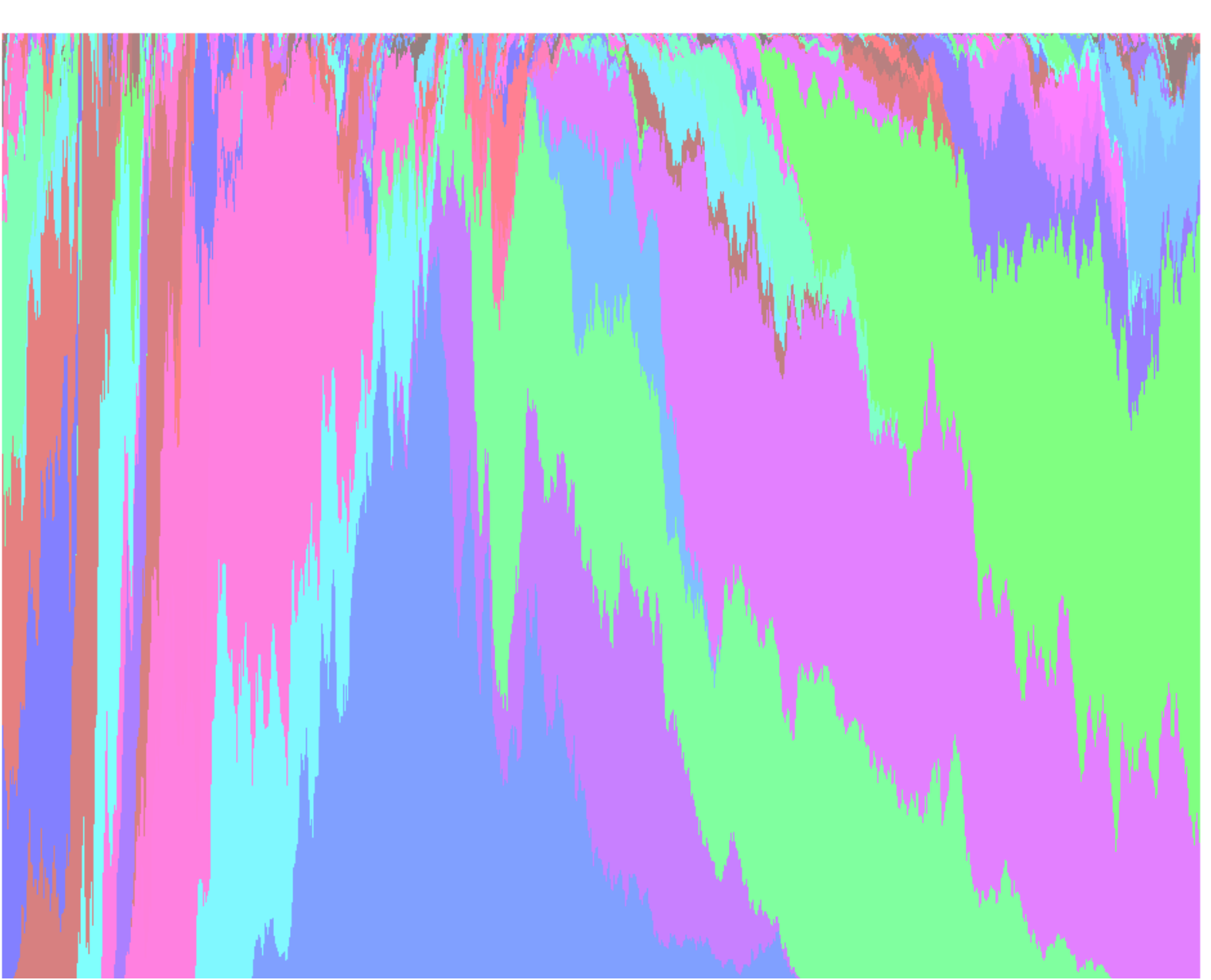}
\end{center}
\caption{Sample path drawn from the Dawson-Watanabe superprocess.  Each colour represents an atom, with height being its (varying) mass.  Left shows $(G_t)$ and right $(G_t/G_t(\XX))$,  a Fleming-Viot process.}\label{super}
\end{figure}

The dynamic model described in the previous section is formulated for discrete time data.  When the time interval between ranking observations is not constant, it is desirable to work with dynamic models  evolving over continuous-time instead, with the underlying random measures $(G_t)$ defined over all $t\in\RR$, but with observations at a discrete set of times $t_1<t_2<\cdots$.  Here we propose a continuous-time model based on the Dawson-Watanabe superprocess \cite{Wat1968a,Daw1975a} (see also \cite{ETHIER1993,Mena2009,Fen2010a,Cox1985}).
This is a diffusion on the space of measures with the gamma process $\Gamma(\alpha,\tau,H)$ as its equilibrium distribution.
It is defined by a generator
\[
\mathcal{L} =  \xi\left( \int G(dX) \frac{\partial^{2}}{\partial
G(X)^{2}} +  \alpha\int H(dX) \frac{\partial
}{\partial G(X)} - \tau \int G(dX) \frac{\partial}{\partial
G(X)} \right)
\]
with $\xi$ parametrizing the rate of evolution.  Figure~\ref{super} gives a sample path, where we see that  it is continuous but non-differentiable.
For efficient inference, it is desirable to be able to integrate out all $G_t$'s except those $G_{t_1},G_{t_2},\ldots$ at observation times. An advantage to using the Dawson-Watanabe superprocess is that, the conditional distribution of $G_{t_s}$ given $G_{t_{s-1}}$ is remarkably simple \cite{EthGri1993a}.  In particular it is simply given by the discrete-time process of the previous section with dependence parameter
$
\phi_{t_s|t_{s-1}} = \frac{\tau}{e^{\tau\xi(t_s-t_{s-1})}-1}.
$
Thus the inference algorithm developed previously is directly applicable to the continuous-time model too.


\section{Experiments}\label{sec:experiments}

\begin{figure}[h]
\begin{center}
\includegraphics[height=.49\textwidth,angle=-90]{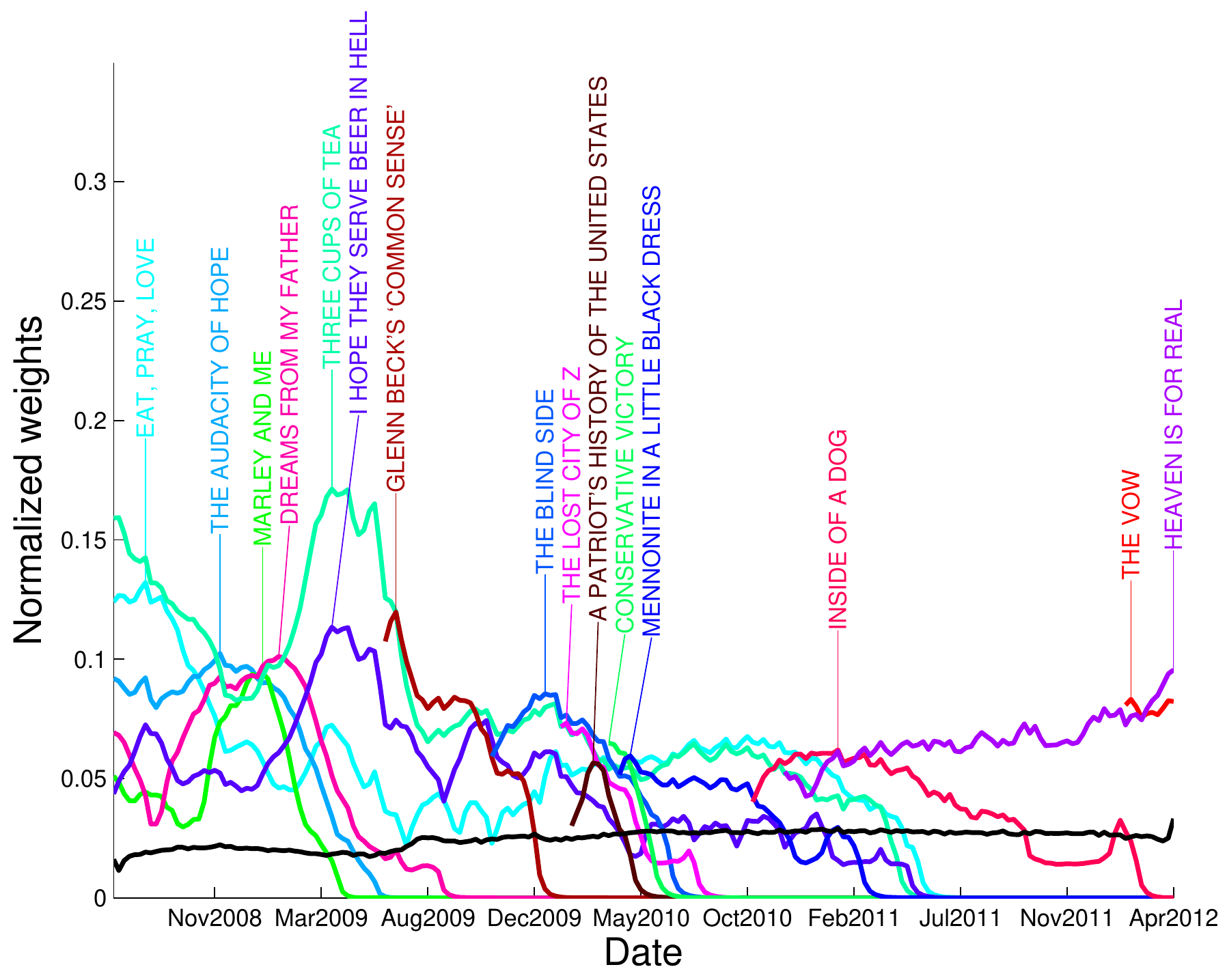}
\includegraphics[height=.49\textwidth,angle=-90]{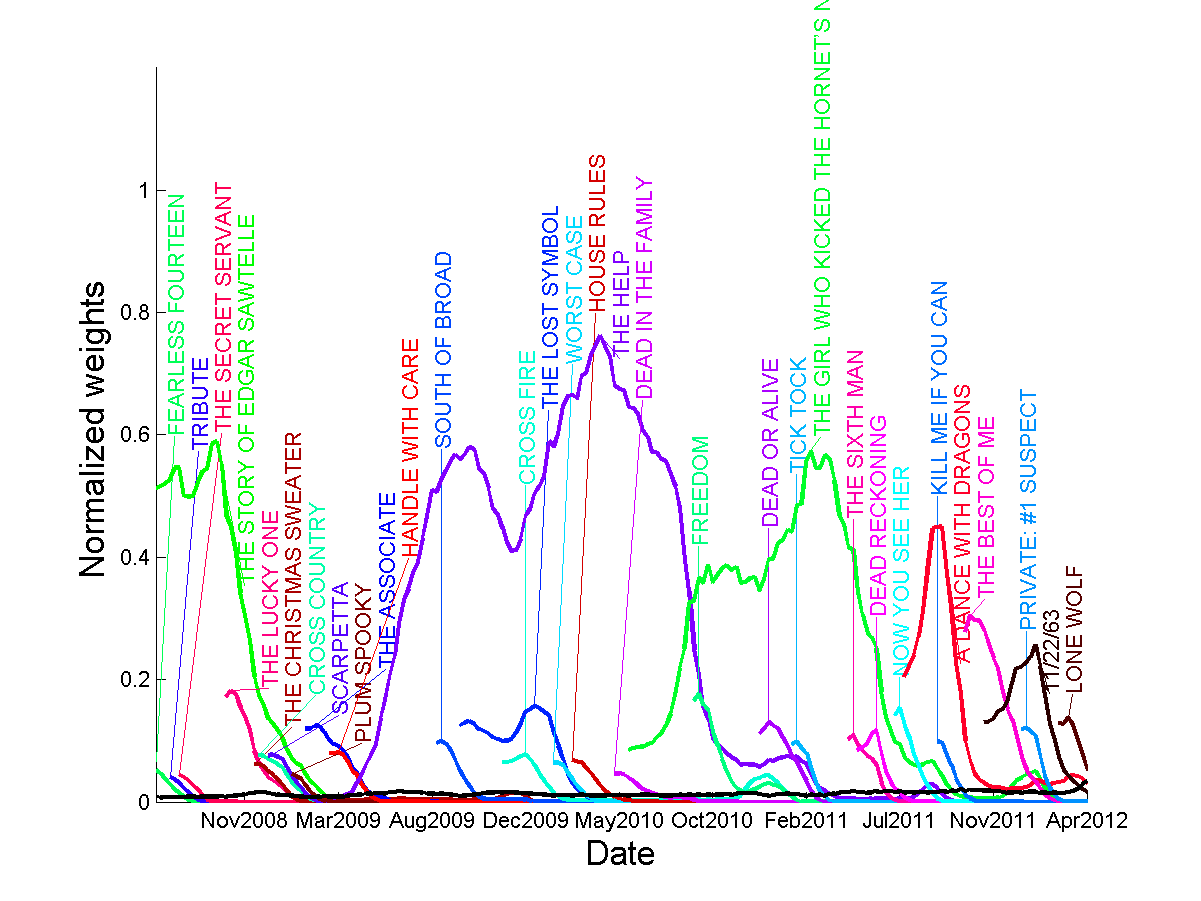}
\end{center}
\caption{Mean normalized weights for paperback nonfiction (left) and hardcover fiction (right). The black lines represent the weight associated to all the books that have not appear in the top-20 lists.}
\label{fig:normweights}
\end{figure}

We apply the discrete-time dynamic Plackett-Luce model to the New York Times bestsellers data. These consist of the weekly top-20 best-sellers list from June 2008 to April 2012 in various categories. We consider here the categories paperback nonfiction (PN) and hardcover fiction (HF), for which respectively 249 and 916 books appear at least once in the top-20 lists over the 200 weeks. We consider that the correlation parameter $\phi_t=\phi$ is constant over time, and assign flat improper priors $p(\alpha)\propto 1/\alpha$ and $p(\phi)\propto 1/\phi$. In order to take into account the publication date of a book, we do not consider books in the likelihood before their first appearance in a list. We run the Gibbs sampler with 10000 burn-in iterations followed by 10000 samples. Mean normalized weights for the more popular books in both categories are shown in Figure~\ref{fig:normweights}.

The model is able to estimate the weights associated to each book that appeared at least once, as well as the total weight associated to all other books, i.e.\ the probability that a new book enters at the first rank in the list, represented by the black curve. Moreover, the Bayesian approach enables us to have a measure of the uncertainty on the weights.
The hardcover fiction category  is characterized by rapid changes in successive lists, compared to the paperback nonfiction. This is quantified by the estimated value of the parameter $\phi$, which are respectively $85\pm 20$ and $140 \pm 40$ for PN and HF. The estimated values of the shape parameter $\alpha$ are $7\pm 1.5$  and $2\pm 1$ respectively.

\section{Discussion}\label{sec:discussion}

We have proposed a Bayesian nonparametric Plackett-Luce model for ranked data.  Our approach is based on the theory of atomic random measures, where we showed that the Plackett-Luce generative model corresponds exactly to a size-biased permutation of the atoms in the random measure.  We characterized the posterior distribution, and derived a simple MCMC sampling algorithm for posterior simulation.  Our approach can be see as a multi-stage generalization of posterior inference in normalized random measures~\cite{James2009,Griffin2011,Favaro2012}, and can be easily extended from gamma processes to general completely random measures.

We also proposed dynamical extensions of our model for both discrete and continuous time data, and applied it to modeling the bestsellers' lists on the New York Times.  Our dynamic extension may be useful for modeling time varying densities or clusterings as well.  In our experiments we found that our model is insufficient to capture the empirical observation that bestsellers often start off high on the lists and tail off afterwards, since our model has continuous sample paths.  We adjusted for this by simply not including books in the model prior to their publication date.  It may be possible to model this better using models with discontinuous sample paths, for example, the Orstein-Uhlenbeck approach of \cite{Gri2011a} where the process evolves via a series of discrete jump events instead of continuously.

\section*{Acknowledgements}

YWT thanks the Gatsby Charitable Foundation for generous funding.

\newpage
\bibliographystyle{unsrt}
\bibliography{ranking}

\appendix

\appendix
\section{Proof of Theorem 1}

The marginal probability \eqref{eq:joint} is obtained by taking the expectation of \eqref{eq:likelihood} with respect to $G$.
Note however that \eqref{eq:likelihood} is a density, so to be totally  precise here we need to work with the \emph{probability} of infinitesimal neighborhoods around the observations instead, which introduces significant notational complexity.  To keep the notation simple, we will work with densities, leaving it to the careful reader to verify that the calculations indeed carry over to the case of probabilities.
\begin{align}
 P((Y_\ell,Z_\ell)_{\ell=1}^L) =& \EE\left[P((Y_\ell,Z_\ell)_{\ell=1}^L|G)\right] \nonumber \\
=& \EE\left[ e^{-G(\mathbb{X})\sum_{\ell i}Z_{\ell i}}
\prod_{k=1}^K G(\{X^*_{k}\})^{n_k}
e^{-G(\{X^*_k\})\sum_{\ell i}(\delta_{\ell ik}-1)Z_{\ell i}}\right] \nonumber \\
\intertext{The gamma prior on $G=\sum_{j=1}^\infty w_j\delta_{X_j}$ is equivalent to a Poisson process prior on $N=\sum_{j=1}^\infty \delta_{(w_j,X_j)}$ defined over the space $\RR^+\times \XX$ with mean intensity $\lambda(w)h(x)$.  Then,}
=& \EE\left[ e^{-\int w N(dw,dx) \sum_{\ell i}Z_{\ell i}}
\prod_{k=1}^K \sum_{j=1}^\infty w_j^{n_k} \bone(X_j=X^*_{k})e^{-w_j\sum_{\ell i}(\delta_{\ell ik}-1)Z_{\ell i}}\right] \nonumber \\
\intertext{Applying the Palm formula for Poisson processes to pull the $k=1$ term out of the expectation,}
=& \int \EE\left[ e^{-\int w (N+\delta_{w^*_1,x^*_1})(dw,dx) \sum_{\ell i}Z_{\ell i}}
\prod_{k=2}^K \sum_{j=1}^\infty w_j^{n_k} \bone(X_j=X^*_{k})
e^{-w_j\sum_{\ell i}(\delta_{\ell ik}-1)Z_{\ell i}}\right]
\nonumber \\&\hspace*{15em}\times
(w_1^*)^{n_k}h(X^*_1) e^{-w_1^*\sum_{\ell i}(\delta_{\ell i1}-1)Z_{\ell i}}\lambda(w^*_1)dw^*_1
\nonumber\\
=&  \EE\left[ e^{-\int w N(dw,dx) \sum_{\ell i}Z_{\ell i}}
\prod_{k=2}^K \sum_{j=1}^\infty w_j^{n_k} \bone(X_j=X^*_{k})
e^{-w_j\sum_{\ell i}(\delta_{\ell ik}-1)Z_{\ell i}}\right]
\nonumber \\&\hspace*{15em}\times
h(X^*_1) \int(w_1^*)^{n_k} e^{-w_1^*\sum_{\ell i}\delta_{\ell i1}Z_{\ell i}}\lambda(w^*_1)dw^*_1
\nonumber\\
\intertext{Now iteratively pull out terms $k=2,\ldots,K$ using the same idea, and we get:}
=&  \EE\left[ e^{-G(\XX) \sum_{\ell i}Z_{\ell i}}
  \right]
\prod_{k=1}^Kh(X^*_k)\int(w_k^*)^{n_k} e^{-w_k^*\sum_{\ell i}\delta_{\ell ik}Z_{\ell i}}\lambda(w^*_k)dw^*_k \nonumber \\
=& e^{-\psi\left(\sum_{\ell i}Z_{\ell i}\right)}
\prod_{k=1}^K h(X^*_k) \kappa\left(n_k,\sum_{\ell i}\delta_{\ell ik}Z_{\ell i}\right)\label{eq:numerator}
\end{align}
This completes the proof of Theorem~\ref{thm:marginal}.

\section{Proof of Theorem 2}

Let $f:\XX\rightarrow \RR$ be measurable with respect to $H$.  Then
the characteristic functional of the posterior $G$ is given by:
\begin{align}
\EE[e^{-\int f(x) G(dx)}|(Y_\ell,Z_\ell)_{\ell=1}^L] =
\frac{\EE[e^{-\int f(x) G(dx)} P((Y_\ell,Z_\ell)_{\ell=1}^L|G)]}{\EE[P((Y_\ell,Z_\ell)_{\ell=1}^L|G)]}\label{eq:charfunc}
\end{align}

The proof is essentially obtained by calculating the numerator and denominator of \eqref{eq:charfunc}.  The denominator is already given in Theorem~\ref{thm:marginal}.  The numerator is obtained using the same technique with the inclusion of the term $e^{\int f(x) G(dx)}$, which gives:
\begin{align}
&\EE\left[e^{-\int f(x) G(dx)} P((Y_\ell,Z_\ell)_{\ell=1}^L|G)\right]\nonumber \\
=& \EE\left[e^{-\int (f(x)+\sum_{\ell i}Z_{\ell i}) G(dx)}\right]
\prod_{k=1}^K h(X^*_k) \int
(w_k^*)^{n_k} e^{-w_k^*(f(X^*_k)+\sum_{\ell i}\delta_{\ell ik}Z_{\ell i})}\lambda(w^*_k)dw^*_k \nonumber \\
\intertext{By the L\'evy-Khintchine Theorem (using the fact that $G$ has a Poisson process representation $N$),}
=& \exp\left(-\int (1-e^{-w(f(x)+\sum_{\ell i}Z_{\ell i})}) \lambda(w)h(x)dwdx\right)\nonumber \\
&\times
\prod_{k=1}^K h(X^*_k) \int
(w_k^*)^{n_k} e^{-w_k^*(f(X^*_k)+\sum_{\ell i}\delta_{\ell ik}Z_{\ell i})}\lambda(w^*_k)dw^*_k \label{eq:denominator}
\end{align}
Dividing the numerator \eqref{eq:numerator} by the denominator \eqref{eq:denominator}, the characteristic functional of the posterior $G$ is:
\begin{align}
&\EE\left[e^{-\int f(x)G(dx)}|(Y_\ell,Z_\ell)_{\ell=1}^L\right] \nonumber\\
=&\exp\left(-\int (1-e^{-wf(x)})e^{-\sum_{\ell i}Z_{\ell i}} \lambda(w)h(x)dwdx\right) \nonumber \\
&\times
\prod_{k=1}^K h(X^*_k) \frac{
\int e^{-f(X^*_k)} (w_k^*)^{n_k} e^{-w_k^*\sum_{\ell i}\delta_{\ell ik}Z_{\ell i}}\lambda(w^*_k)dw^*_k
}{
\int(w_k^*)^{n_k} e^{-w_k^*\sum_{\ell i}\delta_{\ell ik}Z_{\ell i}}\lambda(w^*_k)dw^*_k
}
\end{align}
Since the characteristic functional is the product of $K+1$ terms, we see that the posterior $G$ consists of $K+1$ independent components, one corresponding to the first term above ($G^*$), and the  others corresponding to the $K$ terms in the product over $k$.
Substituting the L\'evy measure $\lambda(w)$ for a gamma process, we note that the first term shows that $G^*$ is a gamma process with updated inverse scale $\tau^*$.  The $k$th term in the product shows that the corresponding component is an atom located at $X^*_k$ with density $(w_k^*)^{n_k} e^{-w_k^*\sum_{\ell i}\delta_{\ell ik}Z_{\ell i}}\lambda(w^*_k)$; this is the density of the gamma distribution over $w^*_k$ in Theorem~\ref{thm:posterior}.  This completes the proof.

\section{Proof of Proposition~\ref{proposition4}}
\label{sec:proofprop4}

We have
\[
P(G_{t}(X_{1k})=0|w_{t-1,,k})=\exp(-\phi_{t-1}w_{t-1,k})
\]
Assume that
\[
P(G_{t}(X_{1k})=0|w_{sk})=\exp(-y_{t|s}w_{sk})
\]
then
\begin{align*}
P(G_{t}(X_{1k})=0|w_{s-1,k})  & =\int\exp(-y_{t|s}w_{sk})p(w_{sk}%
|w_{s-1,k})dw_{sk}\\
& =\sum_{c_{s-1,k}}\int\exp(-y_{t|s}w_{sk})p(w_{sk}|c_{s-1,k})p(c_{s-1,k}%
|w_{s-1,k})dw_{sk}\\
& =\sum_{c_{s-1,k}}\exp\left[  -c_{s-1,k}\log\left(  1+\frac{y_{t|s}}%
{\phi_{s-1}+\tau}\right)  \right]  p(c_{s-1,k}|w_{s-1,k})\\
& =\exp\left(  \frac{-y_{t|s}\phi_{s-1}}{\phi_{s-1}+\tau+y_{t|s}}%
w_{s-1,k}\right)
\end{align*}

\section{Gibbs sampler for the dynamic nonparametric Plackett-Luce model}
\label{sec:gibbstvPL}

For ease of presentation, we assume that $\phi_{t}$ takes the same value
$\phi$ at each time step. The Gibbs sampler will iterate between the following steps

\begin{enumerate}
\item a. For $t=1,\ldots,T$, update $G_{t}(\mathbb{X})$ given $(G_{t-1}%
(\mathbb{X}),\alpha,\phi)$

b. For $t=1,\ldots,T$, update $(c_{t},c_{t\ast})$ given $(w_{t},w_{t\ast
},w_{t+1},w_{t+1\ast},\phi,\alpha)$

\item a. Update $\alpha$ given $(Z,\phi)$

b. For $t=1,\ldots,T$

\qquad Update $w_{t\ast}$ given $(c_{t-1\ast},Z,\phi,\alpha)$

\qquad Update $c_{t\ast}$ given $(w_{t\ast},Z,\phi,\alpha)$

\item For $t=1,\ldots,T$, update ($w_{t},w_{t\ast})$ given $(c_{t-1}%
,c_{t-1\ast},c_{t},c_{t\ast},Z_{t},\alpha,\phi)$

\item For $t=1,\ldots,T$, update $Z_{t}$ given $(w_{t},w_{t\ast})$

\item Update $\phi$ given $w,w_{\ast},\alpha,\phi$\bigskip
\end{enumerate}

The steps are now fully described.\bigskip

\textbf{1.a) Sample }$(G_{t}(\mathbb{X}))$\textbf{ given }$(\alpha,\phi
)$\bigskip

We have
\[
G_{1}(\mathbb{X})|\alpha\sim\text{Gamma}(\alpha,\tau)
\]
and for $t=1,\ldots,T-1$%
\[
G_{t+1}(\mathbb{X})\sim\text{Gamma}(\alpha+M_{t},\tau+\phi)
\]
where $M_{t}\sim$Poisson$(\phi G_{t}(\mathbb{X}))$. The weights $(w_t,w_{t*})$ are then appropriately rescaled. \bigskip

\textbf{1.b) Sample }$(c,c_{\ast})$\textbf{ given (}$w,w_{\ast},\phi,\alpha
)$\bigskip

Consider first the sampling of $c_{1:T}$.\ We have, for $t=1,\ldots,T$ and
$k=1,\ldots,K$
\[
p(c_{tk}|w_{tk},w_{t+1,k})\propto p(c_{tk}|w_{tk})p(w_{t+1,k}|c_{tk})
\]
where
\[
p(c_{tk}|w_{tk})=\text{Poisson}(c_{tk};\phi w_{tk})
\]
and
\[
p(w_{t+1,k}|c_{tk})=\left\{
\begin{array}
[c]{ll}%
\delta_{0}(w_{t+1,k}) & \text{if }w_{tk}=0\\
\text{Gamma}(w_{t+1,k};c_{tk},\tau+\phi) & \text{if }w_{tk}>0
\end{array}
\right.
\]
Hence we can have the following MH update. If $w_{t+1,k}>0$, then we
necessarily have $c_{tk}>0$. We sample $c_{tk}^{\ast}\sim$zPoisson$(\phi
w_{tk})$ where zPoisson$(\phi w_{tk})$ denotes the zero-truncated Poisson
distribution and accept $c_{tk}^{\ast}$ w.p.%
\[
\min\left(  1,\frac{\text{Gamma}(w_{t+1,k};c_{tk}^{\ast},\tau+\phi
)}{\text{Gamma}(w_{t+1,k};c_{tk},\tau+\phi)}\right)
\]

If $w_{t+1,k}=0$, we only have two possible moves: $c_{tk}=0$ or $c_{tk}=1$,
given by the following probabilities%
\begin{align*}
P(c_{tk} &  =0|w_{t+1,k}=0,w_{tk})=\frac{\exp(-\phi w_{tk})}{\exp(-\phi
w_{tk})+\phi w_{tk}\exp(-\phi w_{tk})(\tau+\phi)}=\frac{1}{1+\phi
w_{tk}(\tau+\phi)}\\
P(c_{tk} &  =1|w_{t+1,k}=0,w_{tk})=\frac{\phi w_{tk}\exp(-\phi w_{tk}%
)(\tau+\phi)}{\exp(-\phi w_{tk})+\phi w_{tk}\exp(-\phi w_{tk})(\tau+\phi)}%
=\frac{\phi w_{tk}(\tau+\phi)}{1+\phi w_{tk}(\tau+\phi)}%
\end{align*}

\bigskip

Note that the above Markov chain is not irreducible, as the probability is zero
to go from a state $\left(  c_{tk}>0,w_{t+1,k}>0\right)  $ to a state $\left(
c_{tk}=0,w_{t+1,k}=0\right)  $, even though the posterior probability of this
event is non zero in the case item $k$ does not appear after time $t$. We can
resolve that by the following procedure, that uses a backward forward recursion.

Assume that item $k$ does not appear after time step $\tau_{k}^{+}$ (the same
procedure applies if item $k$ does not appear before time step $\tau_{k}^{-}%
$).\ Then we can sample jointly the whole sequence $(w_{k,t},c_{k,t}%
)_{t=\tau_{k}+1,\ldots,T}$ using the following backward forward recursion.

Let%
\begin{equation}
x_{T}=\sum_{k=1}^{m}Z_{Tk}%
\end{equation}
and for $t=T-1,\ldots,\tau_{k}^+$%
\[
x_{t}=\sum_{k=1}^{m}Z_{tk}+\frac{\phi x_{t+1}}{1+\phi+x_{t+1}}%
\]
\bigskip

We have, for $k=1,\ldots,K$ and $t=\tau_{k}^+$
\begin{align}
c_{tk}|(Z,\phi,w_{tk}) &  \sim\text{Poisson}\left(  \frac{1+\phi}{1+\phi
+x_{t}}\phi w_{tk}\right)  \\
w_{t+1,k}|c_{tk},Z &  \sim\text{Gamma}\left(  c_{k,t},\tau+\phi+x_{t+1}%
\right)
\end{align}\bigskip

\textbf{2.a) Sample }$\alpha$\textbf{ given }$(Z,\phi)$\bigskip

We can sample from the full conditional which is given by
\begin{equation}
\alpha|(Z,\gamma,\phi)\sim\text{Gamma}\left(  a+K,b+y_{1}+\log(1+x_{1}%
)\right)
\end{equation}
where $x_{1}$ and $y_{1}$ are obtained with the following recursion%
\begin{align}
x_{T} &  =\sum_{k=1}^{m}Z_{Tk}\\
y_{T} &  =0
\end{align}
and for $t=T-1,\ldots,1$%
\begin{align*}
x_{t} &  =\sum_{k=1}^{m}Z_{tk}+\frac{\phi x_{t+1}}{1+\phi+x_{t+1}}\\
y_{t} &  =y_{t+1}-\log\left(  \frac{1+\phi}{1+\phi+x_{t+1}}\right)
\end{align*}
$.$\bigskip

\noindent\textbf{2.b) Sample }$(c_{\ast},w_{\ast})$\textbf{ given }%
$(Z,\phi,\alpha)$\bigskip

We can sample from the full conditional which is given by
\begin{equation}
w_{1\ast}|(Z,\phi,\alpha)\sim\text{Gamma}\left(  \alpha,\tau+x_{1}\right)
\end{equation}
where $x_{1}$ is defined above. Then for $t=2,\ldots,T$, let
\begin{align*}
c_{t-1\ast}|(Z,\phi,\alpha,w_{t-1\ast}) &  \sim\text{Poisson}\left(
\frac{1+\phi}{1+\phi+x_{t}}\phi w_{t-1\ast}\right)  \\
w_{t\ast}|c_{t-1\ast},Z,\alpha &  \sim\text{Gamma}\left(  \alpha+c_{t-1\ast
},\tau+\phi+x_{t}\right)
\end{align*}\bigskip

\textbf{3)\ Sample }$(w,w_{\ast})$\textbf{ given }$\left(  Z,\alpha,c,c_{\ast
},\phi\right)  $\bigskip

For each time step $t=1,\ldots,T$

\begin{itemize}
\item For each item $k=1,\ldots,K$, sample
\begin{equation}
w_{tk}|c_{t-1,k},c_{tk},Z_{t}\sim\text{Gamma}\left(  n_{tk}+c_{t-1,k}%
+c_{tk},\tau+2\phi+\sum_{i=1}^{m}\delta_{tik}Z_{ti}\right)
\end{equation}
if $c_{tk}+c_{t-1,k}+n_{tk}>0$, otherwise, set $w_{tk}=0.$ The occurence indicator $\delta_{tik}$ is defined as
\begin{align}
\delta_{t i k} &= \begin{cases}
0 & \text{if $\exists j< i$ with $Y_{t j}=X^*_k$;} \\
1 & \text{otherwise.}
\end{cases}
\end{align}

\item Sample the total mass
\begin{equation}
w_{t\ast}|c_{t\ast},c_{t-1\ast},Z_{t},\alpha\sim\text{Gamma}\left(
\alpha+c_{t\ast}+c_{t-1\ast},\tau+2\phi+\sum_{i=1}^{m}Z_{ti}\right)
\end{equation}
\bigskip
\end{itemize}

\textbf{4) Sample }$Z$\textbf{ given }$(w,w_{\ast})$\bigskip

For $t=1,\ldots,T$ and $i=1,\ldots m$, sample
\begin{equation}
Z_{ti}|w,w_{\ast}\sim\text{Exp}\left(w_{t\ast} + \sum_{k=1}^{K} \delta_{tik}w_{tk}\right)
\end{equation}

\textbf{5) Sample }$\phi$\textbf{ given }$w,w_{\ast},\alpha,\phi$\bigskip

We sample $\phi$ using a MH step. Propose $\widetilde\phi=\phi\exp(\sigma
\varepsilon)$ where $\sigma>0$ and $\varepsilon\sim\mathcal{N}(0,1)$.\ And accept
it with probability
\begin{equation}
\min\left(  1,\frac{p(\widetilde\phi)}{p(\phi)}\frac{\widetilde\phi}{\phi}%
\prod_{t=1}^{T-1}\left [ \frac{p(w_{t+1\ast}|\widetilde\phi,w_{t\ast})}{p(w_{t+1\ast}%
|\phi,w_{t\ast})}\prod_{k=1}^{K}\frac{p(w_{t+1,k}|\widetilde\phi,w_{tk}%
)}{p(w_{t+1,k}|\phi,w_{tk})}\right ]\right)
\end{equation}

\end{document}